\documentclass[letterpaper, 10 pt, conference]{ieeeconf}  

\IEEEoverridecommandlockouts                              

\usepackage[dvipsnames]{xcolor}
\usepackage{amsmath}
\usepackage{amsfonts}
\usepackage{amssymb}
\usepackage{graphicx}
\usepackage{subcaption}
\usepackage{cite}
\let\labelindent\relax
\usepackage{enumitem}
\usepackage{booktabs}
\usepackage{multirow}
\usepackage[hyphens]{url}
\usepackage[hidelinks]{hyperref}
\usepackage{soul,color}
\DeclareSymbolFont{matha}{OML}{txmi}{m}{it}
\DeclareMathSymbol{\varv}{\mathord}{matha}{118}
\DeclareMathOperator{\hyab}{HyAB}
\usepackage{bbm}

\usepackage{soul}

\title{\LARGE \bf
Human-Inspired Topological Representations for Visual Object Recognition in Unseen Environments}

\author{Ekta U. Samani$^{1}$ and Ashis G. Banerjee$^{2}$
\thanks{$^{1}$E. U. Samani was with the Department of Mechanical Engineering, University of Washington, Seattle, WA 98195, USA, during this work
        {\tt\small ektas@uw.edu}}%
\thanks{$^{2}$A. G. Banerjee is with the Department of Industrial \& Systems Engineering and the Department of Mechanical Engineering, University of Washington, Seattle, WA 98195, USA,
        {\tt\small ashisb@uw.edu}}%
}

\begin{document}

\maketitle
\thispagestyle{empty}
\pagestyle{empty}


\begin{abstract}
Visual object recognition in unseen and cluttered indoor environments is a challenging problem for mobile robots.
Toward this goal, we extend our previous work \cite{samani2023persistent} to propose the TOPS2 descriptor, and an accompanying recognition framework, THOR2, inspired by a human reasoning mechanism known as object unity. We interleave color
embeddings obtained using the Mapper algorithm \cite{singh2007topological} for topological soft clustering with the shape-based TOPS descriptor to obtain the TOPS2 descriptor. THOR2, trained using synthetic data, achieves substantially higher recognition accuracy than the shape-based THOR framework and outperforms RGB-D ViT \cite{tziafas2023early} on two real-world datasets: the benchmark OCID dataset and the UW-IS Occluded dataset recorded using commodity hardware. Therefore, THOR2 is a promising step toward achieving robust recognition in low-cost robots.

\end{abstract}



\section{Introduction}

Object recognition is crucial for the semantic understanding of a robot's environment. Early deep learning-based recognition methods are sensitive to environmental variations, making recognizing objects in unseen environments challenging \cite{samani2021visual}. To address such sensitivity, domain adaptation and generalization methods have emerged. Domain adaptation considers a single source and target domain and uses data from the latter for training, making them unsuitable for unknown target domains. Domain generalization methods consider multiple source domains for training, yet the need for abundant real-world training data limits their implementation on robotic systems with commodity hardware \cite{antonik2019human}. 

Alternatively, we consider a single synthetic source domain to obtain object representations suitable for object recognition across multiple target domains. Our previous work \cite{samani2023persistent} proposes a new topological descriptor, TOPS, and an accompanying human-inspired recognition framework, THOR, for recognizing occluded objects in unseen and cluttered indoor environments. THOR shows promising robustness to partial occlusions, but recognition using shape alone is challenging \cite{lai2011large,kasaei2021investigating}. Therefore, multimodal convolutional neural networks \cite{gao2019rgb} and transformer-based approaches \cite{tziafas2023early} have been proposed for RGB-D object recognition. In the absence of large-scale synthetic or real-world RGB-D data (like the RGB-only ImageNet), different approaches for incorporating depth information in learning-based methods have been explored \cite{gao2019rgb,tziafas2023early,gervet2023act3d}. Most of them render depth as a three-channel image for transfer learning-based feature extraction using models pre-trained on RGB images. RGB and depth features are then fused earlier in the model or before the last decision stage. Similarly, classical approaches \cite{browatzki2011going, lai2011large,bo2011depth,bucak2013multiple,paulk2014supervised, fehr2016covariance} compute more than one type of color and shape-based features and fuse them into a global descriptor. 

Following such an approach, in our case, requires object color representations that transfer well from simulation to the real world. However, obtaining them is challenging because the observed chromaticity of objects varies with the lighting conditions \cite{corke2023robotics}. To account for this variation, we follow an approach inspired by the MacAdam ellipses \cite{macadam1942visual} in humans (regions containing indistinguishable colors) to identify the color regions and compute the representations. 
The key contributions of our work are:
\begin{itemize}[leftmargin=*]
    \item We identify color regions (clusters of similar colors) in the standard RGB color space using the Mapper algorithm \cite{singh2007topological} and capture their connectivity in a color network.
    \item We propose a color network-based computation of color embeddings to obtain the TOPS2 descriptor for 3D shape and color-based recognition of occluded objects using an accompanying framework THOR2.
    \item We show that THOR2, trained with synthetic data, outperforms a state-of-the-art transformer 
    adapted for RGB-D object recognition in unseen cluttered environments.
\end{itemize}

\section{Method: THOR2}
\label{ch6method}
\begin{figure*} 
\begin{subfigure}{\textwidth}
  \centering
  \includegraphics[width=0.8\textwidth]{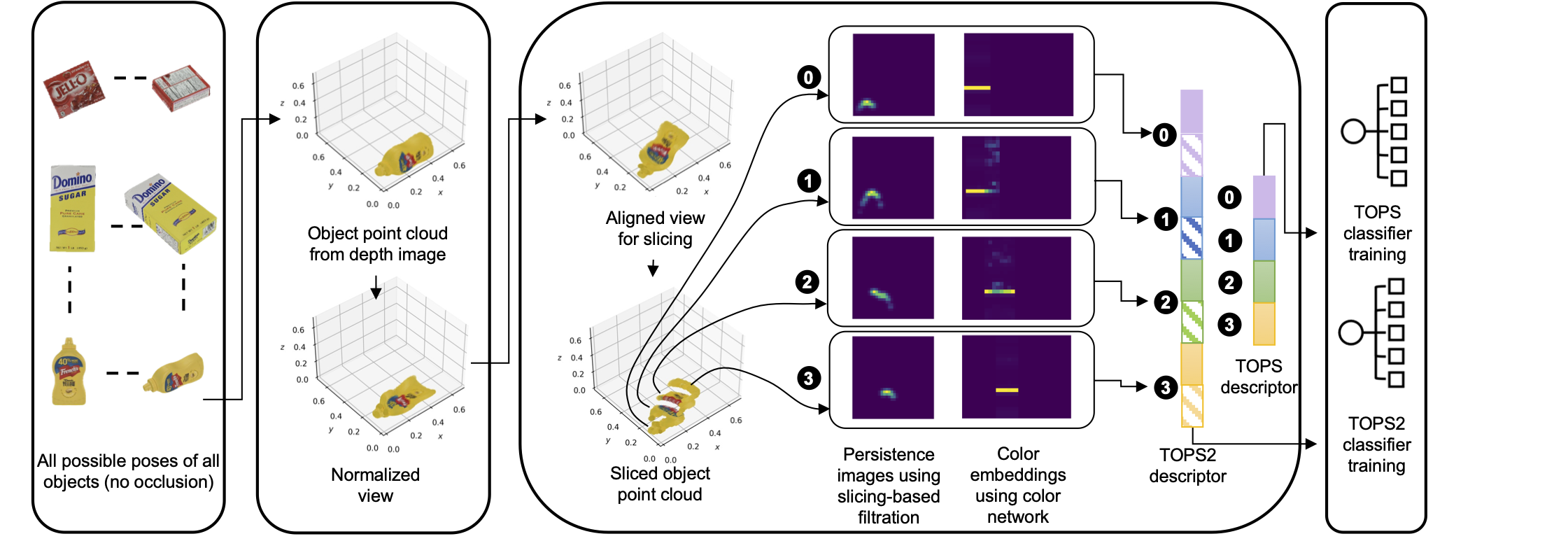} 
  \caption{THOR2 training stage} 
  \label{ch6training}
\end{subfigure}

\begin{subfigure}{\textwidth}
  \centering
  \includegraphics[width=0.8\textwidth]{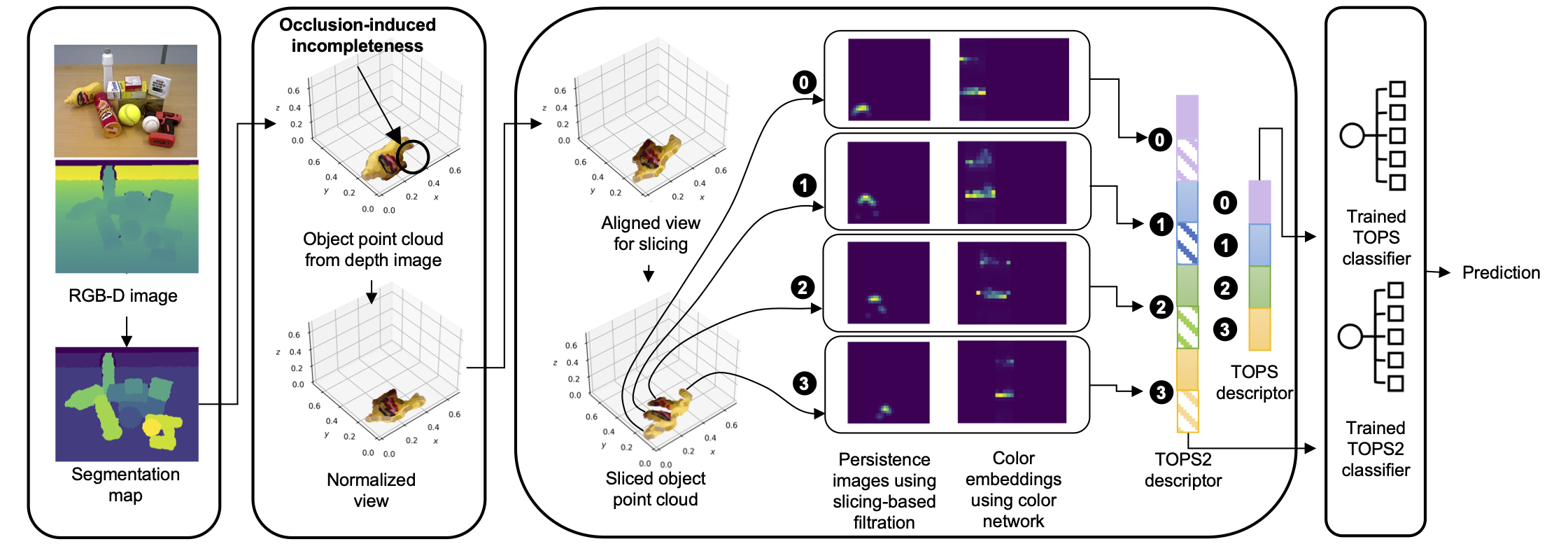}  
  \caption{THOR2 testing stage}
  \label{ch6testing}
\end{subfigure}

\caption{Proposed framework, THOR2, for 3D shape and color-based recognition using object unity \cite{johnson1996perception}, facilitated by the similarity in the TOPS and TOPS2 descriptors of unoccluded and occluded objects.}
\label{ch6pipeline}
\end{figure*}

Given a real-world RGB-D image of an unseen cluttered scene and the corresponding instance segmentation map \cite{xie2021unseen, lu2023self}, our goal is to recognize all the objects in the scene. First, we obtain colored point clouds for every object in the scene and compute the corresponding TOPS and TOPS2 descriptors. The TOPS2 descriptor incorporates color information (in addition to the shape) through embeddings based on the similarity and connectivity among different colors in a color network. The color network is pre-computed using the Mapper algorithm. Last, we perform recognition using two classifiers (one for each descriptor) trained on synthetic RGB-D images. Fig. \ref{ch6pipeline} depicts the overall framework.

\subsection{Color Network Generation}

We consider the colors represented by the standard RGB (sRGB) color space, where values for each color channel range from 0 to 255. Let $X_{rgb}$ denote the set of these colors. The sRGB color space is not perceptually uniform; the Euclidean distance between two colors represented using this color space is not proportional to the difference perceived by humans. Therefore, we convert all the elements in $X_{rgb}$ to the CIELAB color space (also known as the $L^\ast a^\ast b^\ast$ color space) to obtain the set $X_{lab}$. Since the CIELAB color space was designed to capture perceptual uniformity, it is the preferred color space for algorithmically distinguishing between objects by their color \cite{corke2023robotics}.

 We then perform topological soft clustering on $X_{lab}$ using the Mapper algorithm to obtain a color network. Specifically, let $k_{L^\ast}$, $k_{a^\ast}$, and $k_{b^\ast}$ represent the $L^\ast$, $a^\ast$, and $b^\ast$ components of a color $k$ in $X_{lab}$. We use a chroma and hue-based \textit{lens} (i.e., projection) function, $f_l$, to transform the three-dimensional data in $X_{lab}$  to a two-dimensional space. We define $f_l$ as follows.
\begin{equation}
    f_l(k) = \left (\sqrt{k_{a^\ast}^2+k_{b^\ast}^2} , \xi + \arctan \left(\frac{k_{b^\ast}}{k_{a^\ast}}\right) \right ),
\end{equation}

\noindent where $\xi$ is a constant offset selected based on the cover. We adopt the standard choice \cite{chazal2021introduction} of building a cubical \textit{cover} $\mathcal{U}$ by considering a set of regularly spaced intervals of equal length covering the set $f_l(X_{lab})$. Let $r_1$ and $r_2$ (where $r_1, r_2>0$) denote the lengths of the intervals (also known as the \textit{resolution} of the cover) along the two dimensions of $f_l(X_{lab})$, respectively. Let $g_1$ and $g_2$ denote the respective percentages of overlap (also known as the \textit{gain} of the cover) between two consecutive intervals. For each $U \in \mathcal{U}$, a clustering algorithm is applied to $f_l^{-1}(U)$ to obtain the \textit{refined pullback cover} $\mathcal{R}$ of $X_{lab}$. We use the HyAB distance metric \cite{abasi2020distance} to compute the distance between two colors during clustering. Unlike other color difference formulae (e.g., CIEDE2000 \cite{luo2001development}), it applies to a large range of color differences required for practical applications \cite{abasi2020distance}. The HyAB distance between two colors $m$ and $n$ in the CIELAB space is defined as

\begin{equation}
\resizebox{0.95\columnwidth}{!}{%
$\hyab(m,n) = \lvert m_{L^\ast} - n_{L^\ast} \rvert + \sqrt{(m_{a^\ast} - n_{a^\ast})^2 + (m_{b^\ast} - n_{b^\ast})^2 }.$
}
\end{equation}


\begin{figure}[b]
\centering
\includegraphics[width=0.85\columnwidth]{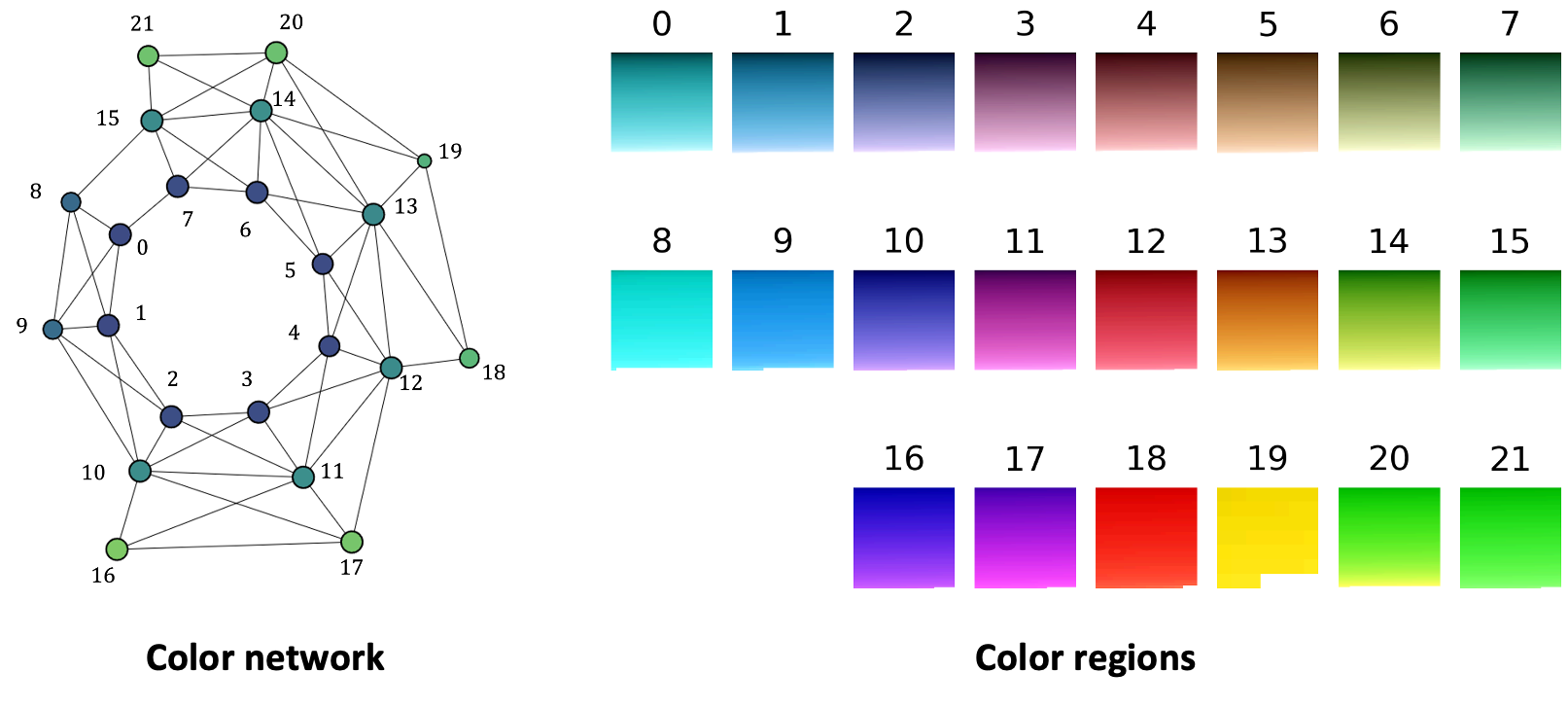}
\caption{The color network that captures the connectivity among color regions identified using the Mapper algorithm.}
\label{mapperoutput}
\end{figure}

\noindent Next, the \textit{nerve} of $\mathcal{R}$ is constructed by collapsing each cluster $R \in \mathcal{R}$ into a vertex and creating a $p$-simplex to represent each $(p+1)$-way intersection of $R$'s. Therefore, in the resulting network, the vertices represent color regions identified by the Mapper algorithm, and the edges represent the overlap between the corresponding color regions. Since a cubical cover does not capture the cyclic nature of the chroma-related dimension of $f_l(X_{lab})$, we add edges connecting the vertices corresponding to the first and last intervals along that dimension. We then eliminate the redundant vertices to obtain the final color network shown in Fig. \ref{mapperoutput}.

Let $G=(V,E)$ be the color network, representing a non-empty set of vertices $V$ and a set of edges $E$. Let $n_c$ represent the number of vertices in $G$, i.e., the number of color regions. We define a similarity matrix, $\Delta$, of size $n_c \times n_c$ to capture the similarity and connectivity between the different color regions in $G$. We define $\Delta$ as follows
\begin{equation}
\Delta = 
\begin{bmatrix}
    \delta_{11} & \delta_{12} &  \dots  & \delta_{1n_c} \\
    \delta_{21} & \delta_{22} &  \dots  & \delta_{2n_c} \\
    \vdots & \vdots & \vdots &  \vdots \\
    \delta_{n_c1} & \delta_{n_c2} & \dots  & \delta_{n_cn_c}
\end{bmatrix}    
\end{equation}
\noindent where $\delta_{i^\prime j^\prime}$ represents the similarity between the $i^\prime$-th and $j^\prime$-th nodes. 
Since every edge in $E$ does not represent the same perceptual difference between the color regions, first, we assign a weight to each edge. Let $\eta_{i^\prime j^\prime}$ be an edge in $E$ that connects the $i^\prime$-th and $j^\prime$-th nodes. To assign the weight, first, we compute the mean color \cite{kasaei2021investigating} of the $i^\prime$-th and $j^\prime$-th colors nodes. The edge $\eta_{i^\prime j^\prime}$ is then assigned a weight equal to the HyAB distance between the mean colors of the $i^\prime$-th and $j^\prime$-th nodes. We then set $\delta_{i^\prime j^\prime} = \frac{1}{1 + l_{i^\prime j^\prime}}$, where $l_{i^\prime j^\prime}$ is the weight of minimum weight path connecting the $i^\prime$-th and $j^\prime$-th nodes. This similarity matrix is pre-computed and used for TOPS2 descriptor computation.


\subsection{TOPS2 Descriptor Computation}

Consider a colored object point cloud $\mathcal{P}$ in $\mathbb{R}^{3}$. Similar to \cite{samani2023persistent}, first, we reorient the point cloud to a reference orientation by performing view normalization. Next, we rotate the view-normalized point $\tilde{\mathcal{P}}$ by an angle $\alpha$ about the $y$-axis to obtain a suitably aligned point cloud $\hat{\mathcal{P}}$. As in \cite{samani2023persistent}, we slice $\hat{\mathcal{P}}$ along the $z$-axis to get slices $\mathcal{S}^i$, where $i \in \mathbb{Z} \cap [0,\frac{h}{\sigma_1}]$. Here, $h$ is the dimension of the axis-aligned bounding box of $\hat{\mathcal{P}}$ along the $z$-axis, and $\sigma_1$ is the thickness of the slices. 
Let $s=(s_x,s_y,s_z)$ represent a point in $\mathcal{S}^i$. For every slice $\mathcal{S}^i$, we modify the $z$-coordinates $\forall s \in \mathcal{S}^i$ to $s_z^\prime$, where $s_z^\prime=i\sigma_1$. 

Next, we perform further slicing of $\mathcal{S}^i$ along the $x$-axis to obtain \textit{strips} $\Omega^j$, where $j \in \mathbb{Z} \cap [0,\frac{w}{\sigma_2}]$. Here, $w$ is the dimension of the axis-aligned bounding box of the slice along the $x$-axis, and $\sigma_2$ represents the 'thickness' of a strip. For every strip $\Omega^j$, we obtain corresponding color vectors $\Phi^j = \big[\begin{smallmatrix} \phi_{1} & \phi_{2} &  \dots  & \phi_{n_c} \end{smallmatrix}\big]^T $ as follows.

\begin{equation}
\phi_\lambda = \sum_{\omega  \: \in \:  \Omega^{j}} \frac{\mathbbm{1}_{X^\lambda_{rgb}}(\omega)}{\sum\limits_{\lambda = 1}^{n_c}\mathbbm{1}_{X^\lambda_{rgb}}(\omega)},
\end{equation}

\noindent where $\lambda \in \{1, \ldots, n_c\}$ represents the $\lambda$-th color region, $\omega$ represents the color of a point in $\Omega^j$ (in the sRGB color space), $\mathbbm{1}$ denotes the indicator function of a set, and $X^\lambda_{rgb}$ represents the set of colors (in the sRGB color space) belonging to the $\lambda$-th color region. Consequently, the color vectors $\Phi^j$ approximately represent the color constitution (in terms of the color regions) of the strips $\Omega^j$.

We then stack the color vectors (with appropriate zero padding) to obtain an $n_s^{max} \times n_c$ dimensional color matrix $\mathcal{C}^i$. Let $\mathcal{C}^i = \big[\begin{smallmatrix}  \mathbf{O}& \ldots & \Phi_{1} & \Phi_{2} & \ldots & \Phi_{n_s} & \ldots & \mathbf{O} \end{smallmatrix}\big]^T $, where $n_s$ is the number of strips in the corresponding slice $\mathcal{S}^i$, $n_s^{max}$ is the maximum number of strips in any given slice, and $\mathbf{O}$  represents a $ n_c \times 1$ dimensional zero matrix. Consequently, the color matrix  $\mathcal{C}^i$ approximately represents the color constitution (in terms of the color regions) of the slice $\mathcal{S}^i$ in a spatially-aware manner. Last, we obtain an embedding $\mathcal{E}^i$ corresponding to the color matrix $\mathcal{C}^i$ as follows.
\begin{equation}
    \mathcal{E}^i = (\mathcal{C}^i\Delta)^T
\end{equation}

\begin{figure}
\centering
\includegraphics[width=0.92\columnwidth]{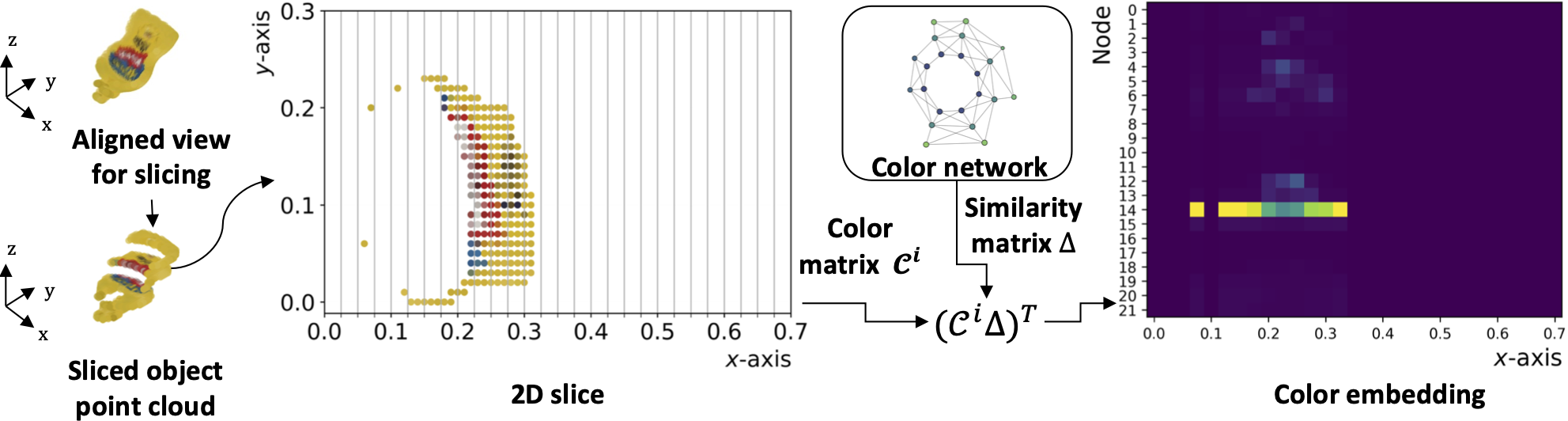}
\caption{Visualization of the color embedding computation for obtaining the TOPS2 descriptor. Example of an aligned object point cloud,  $\hat{\mathcal{P}}$, the slices $\mathcal{S}^0$ to $\mathcal{S}^4$ obtained from it, and the color embedding for one of its slices, i.e., $\mathcal{S}^1$.}
\label{embeddingfigure}
\end{figure}

\noindent Fig. \ref{embeddingfigure} shows this embedding generation for a sample object's slice. We then vectorize the color embeddings and interleave them with the vectorized persistence images from the corresponding TOPS descriptor to obtain the TOPS2 descriptor.

\subsection{THOR2: Training and Testing}
\label{ch6methodtrainingtesting}
Similar to \cite{samani2023persistent}, we consider a training set comprising synthetic RGB-D images corresponding to all the possible views of all the objects. We do not consider any object occlusion scenarios in our training set owing to the slicing-based design of the TOPS2 descriptor; it embodies object unity \cite{johnson1996perception}, enabling an association between the visible part of an occluded object with the original unoccluded object. 
We generate colored object point clouds from the RGB-D images, scale them by a factor of $\sigma_s$, perform view normalization, and 
compute the TOPS and TOPS2 descriptors for them. 
We train one classifier (say $M_{1}$) using the TOPS descriptor and another (say $M_{2}$) using the TOPS2 descriptor.

When testing on a real RGB-D image of a cluttered scene, first, we generate the individual colored point clouds of all the objects using the instance segmentation maps and scale them by a factor of $\sigma_s$. Consider a scaled object point cloud $\mathcal{P}_t$. To recognize $\mathcal{P}_t$, first, we perform view normalization to obtain $\tilde{\mathcal{P}_t}$. Next, we determine if the object corresponding to $\tilde{\mathcal{P}_t}$ is occluded, as described in \cite{samani2023persistent}. If it is occluded, we rotate $\tilde{\mathcal{P}_t}$ by $\pi$ about the $z$-axis to ensure that the first slice on the occluded end of the object is not the first slice during subsequent TOPS and TOPS2 descriptor computation. We then compute the TOPS and TOPS2 descriptors corresponding to $\tilde{\mathcal{P}_t}$ and use the corresponding classifier models, $M_1$ and $M_2$, respectively, to obtain two predictions\footnote{Ablation experiments that use either $M_1$ and $M_2$ alone for recognition achieve 7\% lower accuracy than using both due to a lack of flexibility in selectively using the shape and color information during testing.}. We choose the prediction with the highest probability as our final prediction.



\section{Experiments and Results}
We use synthetic training data from \cite{samani2023persistent} to train THOR2 and evaluate its performance on two real-world datasets: the YCB10 subset of the OCID dataset \cite{suchi2019easylabel} and the UW-IS Occluded dataset \cite{samani2023persistent}. The OCID dataset consists of sequences of increasingly cluttered scenes with up to ten objects recorded by two cameras positioned at different heights and angles. The UW-IS Occluded dataset is recorded using commodity hardware to reflect real-world scenarios with different environmental conditions and degrees of occlusion.

We report performance comparisons with THOR \cite{samani2023persistent} and a Vision Transformer (ViT) adapted for RGB-D object recognition \cite{tziafas2023early}. It uses a surface normal representation \cite{caglayan2022cnns} to render the depth information as a three-channel image and performs \textit{late fusion} of the two modalities. We separately report performance for cases where the RGB-D ViT is trained using synthetic data, and for cases where it is trained using synthetic and real-world data from the YCB dataset \cite{calli2015benchmarking}. We use the Kepler Mapper library \cite{KeplerMapper_JOSS, KeplerMapper_v1.4.1-Zenodo} to compute the color network from $X_{lab}$. We use the CIE standard illuminant D65 for obtaining $X_{lab}$ from $X_{rgb}$. We set $\xi = \frac{\pi}{8}$, and build a cover by choosing $g_1 = 10\%$ and $g_2=25\%$. We set $r_1$ and $r_2$ to divide the corresponding dimensions into three and eight equally-spaced intervals, respectively. For clustering, we use the DBSCAN algorithm \cite{ester1996density}. In the case of THOR and THOR2, we use multi-layer perceptrons for classification and perform training and testing as described in \cite{samani2023persistent}.

\begin{table}[]
\centering
\caption{Comparison of mean recognition accuracy (in \%) on the OCID dataset sequences recorded using cameras placed at different heights.}
\label{ocidtable}
\resizebox{\columnwidth}{!}{%
\begin{tabular}{@{}c|c|cc@{}}
\toprule
\multirow{2}{*}{Method}    & Training        & \multicolumn{2}{c}{Testing data}    \\ \cmidrule(l){3-4} 
                           & data            & Lower camera        & Upper camera        \\ \midrule
THOR                       & Synthetic            & 66.67 $\pm$ 0.22 & 66.50 $\pm$ 0.13 \\
THOR2                      & Synthetic            & \textbf{75.07 $\pm$ 0.33} & \textbf{71.44 $\pm$ 0.15} \\
\midrule
\multirow{3}{*}{RGB-D ViT \cite{tziafas2023early}} & Synthetic            & 56.17 $\pm$ 0.42 & 55.02 $\pm$ 0.23 \\
                           & S + 20\% YCB  & 56.36 $\pm$ 0.31 & 53.73 $\pm$ 0.54 \\
                           & S + 100\% YCB & 57.79 $\pm$ 0.26 & 54.62 $\pm$ 0.19 \\ \bottomrule
\end{tabular}%
}

{\vspace{2mm}\raggedright \underline{Note:} S + \texttt{x}\% YCB indicates that \texttt{x}\% real images from the YCB dataset are used along with the entire synthetic dataset for training and validation. \par}
\end{table}

\begin{figure}
    \centering
    \includegraphics[width=0.85\columnwidth]{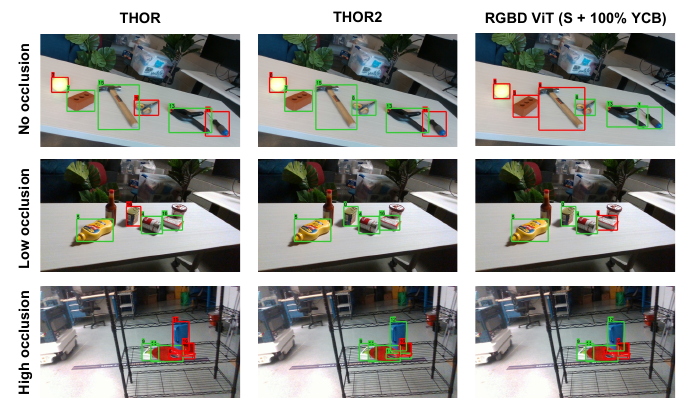}
    \caption{Sample results (green and red boxes indicate correct and incorrect recognition, respectively) from the UW-IS Occluded dataset.}
    \label{ch6uwis2lounge}
\end{figure}

Table \ref{ocidtable} shows that THOR2 achieves higher recognition accuracy than the exclusively shape-based THOR framework on the OCID dataset, regardless of the camera views. Moreover, THOR2 outperforms RGB-D ViT, irrespective of the amount of real-world data used to train it. Similar trends are observed in Table \ref{uwis2finetuning}, where THOR2 achieves substantially higher recognition accuracy than THOR and outperforms RGB-D ViT in all the scenarios of the UW-IS Occluded dataset. Fig. \ref{ch6uwis2lounge} shows a few sample results. It is interesting to note that using additional real-world training data has little impact on RGB-D ViT's performance in the case of the OCID dataset and leads to some performance improvement on the UW-IS Occluded dataset. This observation aligns with our expectations because, unlike the raw and noisy depth images of the UW-IS Occluded dataset, the depth images in the OCID dataset are temporally smoothed. These results also demonstrate that the TOPS and TOPS2 descriptors transfer well to the real world, unlike representations learned using an RGB-D ViT trained on synthetic and limited real-world training data. However, we note that similar to THOR, THOR2 faces difficulty in the case of under-segmentation errors and specific heavy occlusion scenarios \cite{samani2023persistent}. We also implement THOR2 on a LoCoBot equipped with an Intel RealSense D435 camera and an NVIDIA Jetson AGX Xavier processor. THOR2 runs at an average rate of $0.7 s$ per frame in a scene with six objects on this platform.

\begin{table}
\centering
\caption{Comparison of mean recognition accuracy (in \%) in two different environments of the UW-IS Occluded dataset under varying degrees of occlusion.}
\label{uwis2finetuning}
\resizebox{\columnwidth}{!}{%
\begin{tabular}{@{}cc|c|c|ccc@{}}
\toprule
\multirow{2}{*}{Env.} & \multirow{2}{*}{Occlusion} & THOR             & THOR2            & \multicolumn{3}{c}{RGB-D ViT \cite{tziafas2023early}}                                                     \\ \cmidrule(l){3-7} 
                             &                            & Synthetic         & Synthetic         & Synthetic & S+20\% YCB & S+100\% YCB \\ \midrule
\multirow{3}{*}{Warehouse}   & None                       & 51.62 $\pm$ 0.53 & \textbf{61.40 $\pm$ 0.37} & 43.65 $\pm$ 0.70  & 48.14 $\pm$ 1.72   & 49.39 $\pm$ 2.57    \\
                             & Low                        & 48.07 $\pm$ 0.28 & \textbf{58.00 $\pm$ 0.49} & 45.11 $\pm$ 1.37  & 47.56 $\pm$ 1.62   & 47.25 $\pm$ 3.01    \\
                             & High                       & 44.26 $\pm$ 0.25 & \textbf{59.38 $\pm$ 0.35} & 42.52 $\pm$ 1.07  & 48.38 $\pm$ 2.37   & 46.45 $\pm$ 2.87    \\
                             \midrule 
\multirow{3}{*}{Lounge}      & None                       & 56.72 $\pm$ 0.60 & \textbf{64.29 $\pm$ 0.34} & 39.14 $\pm$ 2.07  & 44.72 $\pm$ 2.16   & 46.84 $\pm$ 2.66    \\
                             & Low                        & 54.45 $\pm$ 0.24 & \textbf{65.87 $\pm$ 0.64} & 43.06 $\pm$ 0.60  & 47.41 $\pm$ 0.76   & 47.51 $\pm$ 2.26    \\
                             & High                       & 51.88 $\pm$ 0.46 & \textbf{59.95 $\pm$ 0.52} & 43.50 $\pm$ 1.07  & 46.68 $\pm$ 1.75   & 47.11 $\pm$ 2.90    \\
                             \midrule
\multicolumn{2}{c|}{All}                                   & 52.22 $\pm$ 0.33 & \textbf{62.58 $\pm$ 0.36} & 42.96 $\pm$ 0.87  & 47.04 $\pm$ 1.44   & 47.41 $\pm$ 2.70    \\ \bottomrule
\end{tabular}%
}
\end{table}

\addtolength{\textheight}{-7.5cm} 

\section{Conclusions}
This work presents the TOPS2 descriptor and an accompanying human-inspired recognition framework, THOR2, for 3D shape and color-based recognition of occluded objects in unseen indoor environments. In addition to the persistence images in the TOPS descriptor, TOPS2 comprises color embeddings based on the similarity and connectivity among different colors in a color network obtained using the Mapper algorithm. Our slicing-based approach ensures similarities between the descriptors of the occluded and the corresponding unoccluded objects, facilitating object unity-based recognition. Comparisons on two real-world datasets show that THOR2 benefits from incorporating color information and outperforms RGB-D ViT \cite{tziafas2023early} trained using synthetic and limited real-world data. In the future, we plan to extend THOR2 to incorporate multiple viewpoints to improve the recognition of heavily occluded objects.

\bibliography{references}

\begin{thebibliography}{10}
\providecommand{\url}[1]{#1}
\csname url@samestyle\endcsname
\providecommand{\newblock}{\relax}
\providecommand{\bibinfo}[2]{#2}
\providecommand{\BIBentrySTDinterwordspacing}{\spaceskip=0pt\relax}
\providecommand{\BIBentryALTinterwordstretchfactor}{4}
\providecommand{\BIBentryALTinterwordspacing}{\spaceskip=\fontdimen2\font plus
\BIBentryALTinterwordstretchfactor\fontdimen3\font minus
  \fontdimen4\font\relax}
\providecommand{\BIBforeignlanguage}[2]{{%
\expandafter\ifx\csname l@#1\endcsname\relax
\typeout{** WARNING: IEEEtran.bst: No hyphenation pattern has been}%
\typeout{** loaded for the language `#1'. Using the pattern for}%
\typeout{** the default language instead.}%
\else
\language=\csname l@#1\endcsname
\fi
#2}}
\providecommand{\BIBdecl}{\relax}
\BIBdecl

\bibitem{samani2023persistent}
E.~U. Samani and A.~G. Banerjee, ``Persistent homology meets object unity:
  Object recognition in clutter,'' \emph{IEEE Trans. Robot.}, 2023,
  {C}onditionally accepted for publication, arXiv preprint arXiv:2305.03815.

\bibitem{singh2007topological}
G.~Singh, F.~M{\'e}moli, G.~E. Carlsson \emph{et~al.}, ``Topological methods
  for the analysis of high dimensional data sets and 3{D} object recognition,''
  \emph{PBG@ Eurographics}, vol.~2, pp. 091--100, 2007.

\bibitem{tziafas2023early}
G.~Tziafas and H.~Kasaei, ``Early or late fusion matters: Efficient {RGB-D}
  fusion in vision transformers for 3{D} object recognition,'' in \emph{IEEE
  Int. Conf. Intell. Robot. Syst., To Appear}, 2023, arXiv preprint
  arXiv:2210.00843.

\bibitem{samani2021visual}
E.~U. Samani, X.~Yang, and A.~G. Banerjee, ``Visual object recognition in
  indoor environments using topologically persistent features,'' \emph{IEEE
  Rob. Autom. Lett.}, vol.~6, no.~4, pp. 7509--7516, 2021.

\bibitem{antonik2019human}
P.~Antonik, N.~Marsal, D.~Brunner, and D.~Rontani, ``Human action recognition
  with a large-scale brain-inspired photonic computer,'' \emph{Nat. Mach.
  Intell.}, vol.~1, no.~11, pp. 530--537, 2019.

\bibitem{lai2011large}
K.~Lai, L.~Bo, X.~Ren, and D.~Fox, ``A large-scale hierarchical multi-view
  {RGB-D} object dataset,'' in \emph{IEEE Int. Conf. Rob. Autom.}, 2011, pp.
  1817--1824.

\bibitem{kasaei2021investigating}
S.~H. Kasaei, M.~Ghorbani, J.~Schilperoort, and W.~van~der Rest,
  ``Investigating the importance of shape features, color constancy, color
  spaces, and similarity measures in open-ended 3{D} object recognition,''
  \emph{Intell. Service Rob.}, vol.~14, no.~3, pp. 329--344, 2021.

\bibitem{gao2019rgb}
M.~Gao, J.~Jiang, G.~Zou, V.~John, and Z.~Liu, ``{RGB-D}-based object
  recognition using multimodal convolutional neural networks: a survey,''
  \emph{IEEE Access}, vol.~7, pp. 43\,110--43\,136, 2019.

\bibitem{gervet2023act3d}
T.~Gervet, Z.~Xian, N.~Gkanatsios, and K.~Fragkiadaki, ``Act3{D}: {I}nfinite
  resolution action detection transformer for robotic manipulation,''
  \emph{arXiv preprint arXiv:2306.17817}, 2023.

\bibitem{browatzki2011going}
B.~Browatzki, J.~Fischer, B.~Graf, H.~H. B{\"u}lthoff, and C.~Wallraven,
  ``Going into depth: Evaluating 2{D} and 3{D} cues for object classification
  on a new, large-scale object dataset,'' in \emph{IEEE Int. Conf. Comput. Vis.
  Workshops}, 2011, pp. 1189--1195.

\bibitem{bo2011depth}
L.~Bo, X.~Ren, and D.~Fox, ``Depth kernel descriptors for object recognition,''
  in \emph{IEEE/RSJ Int. Conf. Intell. Robot. Syst.}, 2011, pp. 821--826.

\bibitem{bucak2013multiple}
S.~S. Bucak, R.~Jin, and A.~K. Jain, ``Multiple kernel learning for visual
  object recognition: A review,'' \emph{IEEE Trans. Pattern Anal. Mach.
  Intell.}, vol.~36, no.~7, pp. 1354--1369, 2013.

\bibitem{paulk2014supervised}
D.~Paulk, V.~Metsis, C.~McMurrough, and F.~Makedon, ``A supervised learning
  approach for fast object recognition from {RGB-D} data,'' in \emph{Int. Conf.
  Pervasive Tech. Related Assistive Env.}, 2014, pp. 1--8.

\bibitem{fehr2016covariance}
D.~Fehr, W.~J. Beksi, D.~Zermas, and N.~Papanikolopoulos, ``Covariance based
  point cloud descriptors for object detection and recognition,'' \emph{Comput.
  Vis. Image Understanding}, vol. 142, pp. 80--93, 2016.

\bibitem{corke2023robotics}
P.~Corke, \emph{Robotics, Vision and Control: Fundamental Algorithms in
  Python}.\hskip 1em plus 0.5em minus 0.4em\relax Springer Nature, 2023, vol.
  146.

\bibitem{macadam1942visual}
D.~L. MacAdam, ``Visual sensitivities to color differences in daylight,''
  \emph{Josa}, vol.~32, no.~5, pp. 247--274, 1942.

\bibitem{xie2021unseen}
C.~Xie, Y.~Xiang, A.~Mousavian, and D.~Fox, ``Unseen object instance
  segmentation for robotic environments,'' \emph{IEEE Trans. Rob.}, vol.~37,
  no.~5, pp. 1343--1359, 2021.

\bibitem{lu2023self}
Y.~Lu, N.~Khargonkar, Z.~Xu, C.~Averill, K.~Palanisamy, K.~Hang, Y.~Guo,
  N.~Ruozzi, and Y.~Xiang, ``Self-supervised unseen object instance
  segmentation via long-term robot interaction,'' \emph{arXiv preprint
  arXiv:2302.03793}, 2023.

\bibitem{johnson1996perception}
S.~P. Johnson and R.~N. Aslin, ``Perception of object unity in young infants:
  {T}he roles of motion, depth, and orientation,'' \emph{Cognitive
  Development}, vol.~11, no.~2, pp. 161--180, 1996.

\bibitem{chazal2021introduction}
F.~Chazal and B.~Michel, ``An introduction to topological data analysis:
  {F}undamental and practical aspects for data scientists,'' \emph{Frontiers
  Artificial Intell.}, vol.~4, p. 667963, 2021.

\bibitem{abasi2020distance}
S.~Abasi, M.~Amani~Tehran, and M.~D. Fairchild, ``Distance metrics for very
  large color differences,'' \emph{Color Research \& Appl.}, vol.~45, no.~2,
  pp. 208--223, 2020.

\bibitem{luo2001development}
M.~R. Luo, G.~Cui, and B.~Rigg, ``The development of the {CIE} 2000
  colour-difference formula: {CIEDE}2000,'' \emph{Color Research \&
  Application: Endorsed by Inter-Society Color Council, The Colour Group (Great
  Britain), Canadian Society for Color, Color Science Association of Japan,
  Dutch Society for the Study of Color, The Swedish Colour Centre Foundation,
  Colour Society of Australia, Centre Fran{\c{c}}ais de la Couleur}, vol.~26,
  no.~5, pp. 340--350, 2001.

\bibitem{suchi2019easylabel}
M.~Suchi, T.~Patten, D.~Fischinger, and M.~Vincze, ``Easy{L}abel: {A}
  semi-automatic pixel-wise object annotation tool for creating robotic {RGB-D}
  datasets,'' in \emph{IEEE Int. Conf. Rob. Autom.}, 2019, pp. 6678--6684.

\bibitem{caglayan2022cnns}
A.~Caglayan, N.~Imamoglu, A.~B. Can, and R.~Nakamura, ``When {CNN}s meet random
  {RNN}s: Towards multi-level analysis for {RGB-D} object and scene
  recognition,'' \emph{Comput. Vis. Image Understanding}, vol. 217, p. 103373,
  2022.

\bibitem{calli2015benchmarking}
B.~Calli, A.~Walsman, A.~Singh, S.~Srinivasa, P.~Abbeel, and A.~M. Dollar,
  ``Benchmarking in manipulation research: Using the {Y}ale-{CMU}-{B}erkeley
  object and model set,'' \emph{IEEE Rob. Autom. Mag.}, vol.~22, no.~3, pp.
  36--52, 2015.

\bibitem{KeplerMapper_JOSS}
\BIBentryALTinterwordspacing
H.~J. van Veen, N.~Saul, D.~Eargle, and S.~W. Mangham, ``Kepler mapper: A
  flexible python implementation of the mapper algorithm.'' \emph{J. Open
  Source Software}, vol.~4, no.~42, p. 1315, 2019. [Online]. Available:
  \url{https://doi.org/10.21105/joss.01315}
\BIBentrySTDinterwordspacing

\bibitem{KeplerMapper_v1.4.1-Zenodo}
\BIBentryALTinterwordspacing
------, ``{Kepler {M}apper: {A} flexible {P}ython implementation of the
  {M}apper algorithm},'' Oct. 2020. [Online]. Available:
  \url{https://doi.org/10.5281/zenodo.4077395}
\BIBentrySTDinterwordspacing

\bibitem{ester1996density}
M.~Ester, H.-P. Kriegel, J.~Sander, X.~Xu \emph{et~al.}, ``A density-based
  algorithm for discovering clusters in large spatial databases with noise,''
  in \emph{Int. Conf. Knowl. Discovery Data Mining}, vol.~96, no.~34, 1996, pp.
  226--231.

\end{thebibliography}

\bibliographystyle{IEEEtran}

\end{document}